\documentclass[10pt]{article} 
\usepackage[preprint]{rlc}

\usepackage{amssymb}            
\usepackage{amsmath}
\usepackage{amsthm}
\usepackage{mathtools}          
\usepackage{mathrsfs}           
\mathtoolsset{showonlyrefs}     
\usepackage{graphicx}           
\usepackage{subcaption}         
\usepackage[space]{grffile}     
\usepackage{url}                
\usepackage[multiple]{footmisc}
\usepackage[normalem]{ulem}
\usepackage{enumitem}
\usepackage{multirow}
\usepackage{natbib}
\usepackage{etoolbox}
\usepackage{rotating}
\usepackage{soul}
\usepackage{xspace}

\graphicspath{{figs/}}

\newcommand{\expecb}[1]{\mathbb{E}\Big[#1\Big]}

\DeclareMathOperator*{\argmax}{arg\,max}

\newcommand{\nStates}{$716$\xspace}
\newcommand{\nStatesMinusOne}{$715$\xspace}

\title{ICU-Sepsis: A Benchmark MDP Built from Real Medical Data}

\author{Kartik Choudhary, Dhawal Gupta, and Philip S. Thomas \\
    \{kartikchoudh,dgupta,pthomas\}@cs.umass.edu \\
    College of Information and Computer Sciences \\
    University of Massachusetts
    }

\begin{document}

\maketitle

\begin{abstract}

We present \emph{ICU-Sepsis}, an environment that can be used in benchmarks for evaluating reinforcement learning (RL) algorithms. Sepsis management is a complex task that has been an important topic in applied RL research in recent years. Therefore, MDPs that model sepsis management can serve as part of a benchmark to evaluate RL algorithms on a challenging real-world problem.
However, creating usable MDPs that simulate sepsis care in the ICU remains a challenge due to the complexities involved in acquiring and processing patient data.
ICU-Sepsis is a lightweight environment that models personalized care of sepsis patients in the ICU. The environment is a tabular MDP that is widely compatible and is challenging even for state-of-the-art RL algorithms, making it a valuable tool for benchmarking their performance.
However, we emphasize that while ICU-Sepsis provides a standardized environment for evaluating RL algorithms, it should not be used to draw conclusions that guide medical practice.
\end{abstract}


\section{Introduction}\label{sec:intro}

In this paper, we present \emph{ICU-Sepsis}---an easy-to-use environment that can be used in benchmarks for \emph{reinforcement learning} (RL) algorithms. This environment is a \emph{Markov decision process} (MDP) that models the problem of providing personalized care to sepsis patients, constructed using real-world medical records. 
The environment exhibits a level of complexity that challenges state-of-the-art RL algorithms, making it a suitable domain to include when benchmarking and evaluating RL algorithms. 
Its tabular nature makes it a lightweight and portable MDP that is compatible with many RL algorithms and which can be quickly incorporated into any benchmark suite. 

Sepsis is a life-threatening condition that arises when the body's response to infection causes injury to its own tissues and organs, and requires personalized care based on a sequence of clinical decisions. This sequence of decisions results in evaluative feedback---information about whether or not the patient survived. 
%
However, this feedback does not specify what the optimal decisions would have been in retrospect, i.e., it does not provide the instructive feedback required for supervised learning (e.g., what the optimal dosages of each medicine would have been). The evaluative nature of this feedback and the potential for delays in its availability make reinforcement learning methods a natural choice for this problem. 

Following the work of \citet{komorowski2018the}, sepsis management has emerged as a prominent use case in applied RL research \citep{raghu2019reinforcement, Yu2023}, where historical patient data obtained from large medical record databases is used to model sepsis as an MDP.  
One of the most common sources of patient records is the MIMIC-III database \citep{mimic3}, which contains health-related data for over forty thousand ICU patients, collected between 2001 and 2012.
Recognizing the widespread interest and importance of this topic, a dedicated RL environment that emulates the environments used in applied RL research for sepsis treatment in the ICU can serve as a valuable tool for evaluating the efficacy of RL algorithms for a real-world problem of interest. 

\begin{figure}[h]
\centering
    \includegraphics[width=\textwidth]{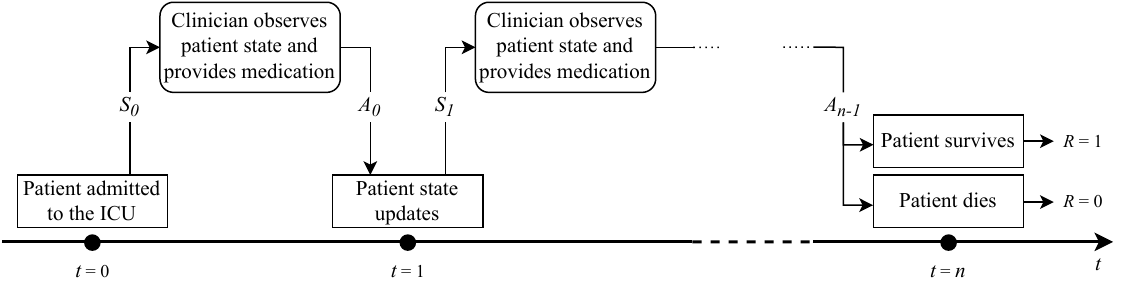}
\caption{Illustration of one episode in the ICU-Sepsis environment. The clinician treats the patient through \emph{actions}, which affect how their \emph{state} evolves over time, until the patient is discharged (and a positive reward is received), or the patient dies (and no reward is received).}
\label{fig:mdp-episode-illustration2}
\end{figure}

Various researchers have developed MDPs that simulate sepsis, as described in detail in Section \ref{subsec:sepsis-in-rl}.
However, constructing such an MDP is a complex process of querying, cleaning, and filtering patient data from a medical database.
Slight differences in the design and implementation of these procedures by different researchers have resulted in slightly different MDPs. 
Consequently, a standardized version of the sepsis MDP, essential for establishing a benchmark, has yet to be defined.
Moreover, although the MIMIC-III database is openly available, researchers must formally request access, a process that entails completing a data protection course and signing a data use agreement. While these measures are crucial for upholding patient privacy, they, in conjunction with the complex and varying MDP creation processes, pose significant challenges for RL researchers seeking to include sepsis treatment in their benchmark suites.

ICU-Sepsis addresses these issues by presenting users with a readily deployable environment, designed for evaluating the efficacy of most RL algorithms. The MDP is a standalone environment built with the MIMIC-III database that does not require any querying, cleaning, or filtering from the user and can be used or modified without restriction (i.e., users need not complete courses or sign a data use agreement) while maintaining patient privacy (see Section \ref{subsec:privacy} for details). 

Following the precedent set 
by \citet{komorowski2018the}, the status of a patient at any given time is discretized into a set of \nStates states,\footnote{\citet{komorowski2018the} constructed an MDP with \textit{roughly} 750 states. After removing some problematic states (as discussed later), and introducing additional states to model termination, the ICU-Sepsis MDP that we present contains \nStates states.} balancing the granularity of the state set with the amount of data available for modeling each state transition probability. 
Similarly, following prior work \citep{komorowski2018the}, the possible medical interventions by clinicians are discretized into 25 possible actions.
The discount factor $\gamma$ is set to $1$ to reflect the goal of maximizing each patient's chance of survival. At the end of each episode, patient survival results in a reward of \(+1\), while death corresponds to a reward of \(0\), with all intermediate rewards also being \(0\). 
Figure \ref{fig:mdp-episode-illustration2} shows an illustration of one episode in the ICU-Sepsis environment.
An agent selecting actions uniformly randomly achieves an expected return (probability of patient survival) of \(0.78\), while an
optimal policy computed using value iteration \citep{Bellman:1957} achieves an expected return of \(0.88\).

The ICU-Sepsis MDP is provided in a GitHub repository.\footnote{\url{https://github.com/icu-sepsis/icu-sepsis}}
To allow researchers to quickly implement the environment in the software of their choice, the environment is provided as a set of CSV files containing the transition, reward, and initial state distribution matrices, as well as open-source Python implementations in OpenAI Gym \citep{brockman2016openai} and Gymnasium \citep{towers2023gymnasium}.  
%
 See Section \ref{sec:software} for details.

\section{Background}\label{sec:background}

In this section we present the notation and terminology that we use for RL, provide background regarding sepsis management, and review prior work that models sepsis treatment as an RL problem.

\subsection{Technical setting}

RL problems are often modeled as an agent interacting with a discrete-time Markov decision process (MDP) \citep{sutton2018rlbook,Frnkranz2011}. 
Formally, an MDP is a tuple of the form $(\mathcal S, \mathcal A, p, R, d_0)$,
where the state set \(\mathcal S\) contains all possible states of the environment, and the set of actions available to the agent in state $s \in \mathcal S$ is denoted by $\mathcal A(s)$. The set of all possible actions in any state is denoted by
\vspace{-2pt}
\begin{equation}
    \mathcal{A}^+ \doteq \bigcup_{s\in\mathcal S} \mathcal A(s).
\end{equation}
In this work we consider MDPs where $\mathcal A^+$ and $\mathcal S$ are finite, unless stated otherwise.
The transition function \(p: \mathcal{S} \times \mathcal{A}^+ \times \mathcal{S} \to [0,1] \) defines the probabilities of transitioning from one state to the next after taking an action: \(p(s, a, s') \doteq \Pr(S_{t+1}{=}s' | S_t{=}s, A_t{=}a)\). 
The function \(R: \mathcal{S} \times \mathcal{A}^+ \times \mathcal{S} \to [0,1]\) gives the reward when transitioning from one state to another after taking an action. In general, this reward can be stochastic, but in our case, it is a deterministic function of \(S_t, A_t\) and \(S_{t+1}\), written as \(R_t = R(S_t, A_t, S_{t+1})\).
The initial-state distribution function \(d_0: \mathcal{S} \to [0,1]\) characterizes the distribution of the initial state: \(d_0(s) \doteq \Pr(S_0=s)\).

At any given integer time \(t \geq 0\), the agent is in a state \(S_t \in \mathcal S\), and the agent-environment interaction takes place by the agent taking action \(A_t \in \mathcal A(S_t)\), transitioning to the next state \(S_{t+1} \sim p(S_t, A_t, \cdot)\), and receiving a reward \(R_t = R(S_t, A_t, S_{t+1})\). 
A policy \(\pi: \mathcal{S} \times \mathcal{A}^+ \to [0,1]\) defines the probability of taking each action given a state: \(\pi(s, a) \doteq \Pr(A_t{=}a | S_t{=}s)\).
A trajectory \(H\) of length \(L\) can be defined as a sequence of \(L\) (state, action, reward) tuples: \(H \doteq (S_0, A_0, R_0, S_1, \dots, S_{L-1}, A_{L-1}, R_{L-1})\). A dataset \(D\) is defined as a collection of such trajectories: \(D \doteq \{H^{(0)}, H^{(1)}, \dots, H^{(N-1)}\}\).

The \textit{return} of a trajectory is the discounted sum of rewards \(G(H) \doteq \sum_{t=0}^\infty \gamma^t R_t\), where \(\gamma \in [0,1]\) is the discount factor that determines the relative weight of future and immediate rewards.
The \textit{objective function} \(J(\pi)\) is the performance measure of a policy \(\pi\), defined as the expected return when the agent uses the policy $\pi$ to select actions: \(J(\pi) \doteq \expecb{\sum_{t=0}^\infty \gamma^t R_t}.\)
%
The goal of an RL agent is to find an optimal policy \(\pi^*\), which is a policy that maximizes the expected return: \(\pi^* \in \argmax_{\pi} J(\pi)\).  
%

\subsection{Sepsis management}

Sepsis is a life-threatening organ dysfunction caused by a dysregulated host response to infection \citep{Singer2016}, and is implicated in approximately 1 in every 5 deaths worldwide \citep{Rudd2020Global}. 
It is a severe multisystem disease with high mortality rates, and it is challenging to determine the correct treatment strategy for its various manifestations \citep{Polat2017}.

Sepsis management is a sequential decision-making problem, wherein clinicians make a series of medical interventions based on the state of the patients, to provide treatments that maximize the chances of patient survival. 
Guidelines such as those published by the Surviving Sepsis Campaign \citep{Evans2021} provide valuable frameworks for early recognition and key interventions.
However, owing to the complex nature of the condition, there are ongoing efforts to further refine guidelines and individualize treatment approaches \citep{Kissoon2014, Kalil2017Infectious}.
In the event of a patient's death, it is generally not possible to determine the precise steps in their care that, if changed, would have resulted in their survival. Likewise, figuring out how to modify policies to enhance survival prospects for future patients remains an ongoing and critical challenge.

%

\subsection{RL for sepsis treatment}\label{subsec:sepsis-in-rl}

There has been significant interest recently in the healthcare domain in using historical patient data to learn new policies for patient care, such as for diabetes \citep{bastani2014model}, epilepsy treatment \citep{pineau2009treating}, cancer trials \citep{Humphrey2017UsingRL}, radiation adaptation for lung cancer \citep{Tseng2017}, and many others as shown by \citet{yu2020reinforcement}. 
In the context of sepsis management, datasets like MIMIC-III \citep{mimic3} and e-ICU \citep{Pollard2018eicu} have been used to create tabular MDPs to find better treatment methods for sepsis \citep{komorowski2018the, oberst2019counterfactual, Tsoukalas2015, pittir38498lyu} and more specialized cases, such as pneumonia-related sepsis \citep{pittir8143kreke}, as well as optimizing the initial response to sepsis \citep{Rosenstrom2022}.
\citet{Nanayakkara2022} combined distributional deep reinforcement learning \citep{bdr2023} with mechanistic physiological models \citep{Hodgkin1952, Bezzo2018} to devise personalized sepsis treatment strategies.
\citet{DBLP:journals/corr/abs-1711-09602} studied the use of deep reinforcement learning with continuous states for optimizing sepsis treatments. 

RL researchers may want to ensure that the algorithms that they develop are effective for important real-world problems like sepsis treatment. 
However, different (but similar) environment models are used in the applied RL research described above, and recreating these environment models can be challenging. 
Our work therefore seeks to provide a standardized RL environment that simulates sepsis treatment in the ICU. 
This environment is designed to be an easy-to-use environment within RL algorithm benchmarks, which is also representative of an important real problem. 
Although ICU-Sepsis is built from real data, and follows procedures from prior work intended to guide medical practice, the environment that we present is only intended for use as a standardized MDP to evaluate RL algorithms, not as a tool for studying sepsis treatment or guiding medical practice.

\section{Software and Data}\label{sec:software}

The dynamics of the ICU-Sepsis environment are available to download as \texttt{.csv} tables from the GitHub repository.\footnote{\url{https://github.com/icu-sepsis/icu-sepsis}} The use of \texttt{.csv} files allows for development with different libraries and programming languages.
We also provide Python code compatible with the widely-used frameworks OpenAI Gym \citep{brockman2016openai} and Gymnasium \citep{towers2023gymnasium}.

\subsection{The environment parameters and implementation }

The states $\mathcal S = \{0, 1, \dots, \nStatesMinusOne \}$ and actions $\mathcal A^+ = \{0, 1, \dots, 24 \}$ are both represented by integers. 
The transition tensor table has \(|\mathcal S|\times |\mathcal A^+| = 17,\!900\) rows and \(|\mathcal S|\) columns. The value \(p(s,a,s')\) is present in the \((s')^\text{th}\) column of the \((s\cdot |\mathcal A^+| + a)^\text{th}\) row. The centroids of the state clusters are provided in an optional table that has $|\mathcal S|$ rows and 47 columns, with the $s^\text{th}$ row containing the 47-dimensional centroid of state $s$ in the normalized feature space.

The table representing the initial state distribution as a vector has 1 row and \(|\mathcal S|\) columns. The value of \(d_0(s)\) is present in the \(s^\text{th}\) column.
The reward table also has 1 row and \(|\mathcal S|\) columns, with the value of \(R(s,a,s')\) present in the \((s')^\text{th}\) column. Details of reproducing these parameters from the MIMIC-III dataset are given in Appendix \ref{adx:reproducing}.

\section{The ICU-Sepsis Environment}\label{sec:env_desc}



Hospitals systematically monitor various patient statistics and vitals, documented in their \textit{electronic health records} (EHRs) \citep{Shvo2017}, during the course of patient care.
%
%
Clinicians prescribe appropriate medication using the collected data, adjusting dosages as the patient's condition evolves.
In recent years, a growing number of hospitals have taken to recording detailed patient treatment information within their EHR systems.
This rich dataset allows for the extraction of valuable insights, enabling the development of informed policies geared towards enhancing patient care.

\subsection{Formulating sepsis management as a reinforcement learning problem}\label{subsec:problem_as_rl}

Based on the statistics collected by the hospital, at any given point in time, a patient's health can be described by a vector representing different features of the patient, such as their demography, vitals, body fluid levels, etc. 
After discretizing time into uniform chunks, these features can be clustered into a finite set \(\mathcal S\), thus representing the evolution of the status of a patient in the hospital as a sequence of discrete states across discrete time steps.
The different types and dosages of medications administered to the patient can similarly be represented as a finite set of discrete actions $\mathcal A^+$.
The number of different medications \(d_A\) and
number of dosage levels \(n_A\) of each
%
%
medication determines the size of the action set: \(|\mathcal A^+| = (n_A)^{d_A}\).

The EHR data for \(|D|\) patients can be represented as a dataset $D$, where each trajectory describes the hospitalization of one patient. The reward associated with each time step is \(R=0\), except for the last time step, where the reward is \(R=+1\) if the patient survives. This design choice causes the expected return to correspond to the probability of a randomly selected patient surviving.

\subsection{The ICU-Sepsis dataset}\label{subsec:icu-dataset}

The dataset $D$ is created by using real patient data describing approximately 17,000 sepsis patients from version 1.4 of the MIMIC-III dataset \citep{mimic3}.
Following the procedure by \citet{komorowski2018the}, time is discretized into 4-hour blocks, and the states are clustered using the K-means clustering algorithm \citep{macqueen1967some} with K-means++ initialization \citep{David2007}.
Three additional states are added to model termination---two corresponding to survival and death, based on 90-day mortality, and the third as the \emph{terminal absorbing state} \(s_\infty\).
Actions specify the dosages of intravenous fluids and vasopressors (two different interventions) with similar discretization thresholds as used by \citet{komorowski2018the}.

In many states, not all actions are seen enough times to enable accurate estimation of the transition probabilities $p(s,a,\cdot)$. 
Therefore, for any given state-action pair \((s, a)\), the action \(a\) is considered an admissible action for state \(s\) if and only if it occurs at least \(\tau\) times in state \(s\) within the dataset, and the parameter \(\tau\) is called the \textit{transition threshold}.
The set of all such admissible actions for any given state \(s\) is denoted by \(\mathcal A(s) \subseteq \mathcal A^+\).
Based on this definition of admissible actions, some states have no admissible actions at all, and such states are removed from the MDP.

\subsection{Constructing the ICU-Sepsis MDP}\label{subsec:const-mdp}


Given a dataset \(D\) of trajectories, the indicator for state-action-next-state tuple \((s,a,s')\) at time-step \(t\) in trajectory \(h\) is given by

\vspace{-18pt}

\begin{equation}
    I_D (h, t, s, a, s') \doteq \begin{cases}
        \begin{aligned}
            1 \qquad &\text{if } s{=}S_t^{(h)}, a{=}A_t^{(h)}, s'{=}S_{t+1}^{(h)}  \\
            0 \qquad &\text{otherwise,}
        \end{aligned}
    \end{cases}
\end{equation}
for $s,s' \in \mathcal S^2, a \in \mathcal A^+, t\in \{0, 1, \dots, \},$ and $h\in D$. 
This indicator is used to define the set of admissible actions \(\mathcal A(s)\) in a given state \(s\in\mathcal S\) as

\vspace{-20pt}

\begin{equation}
    \mathcal A(s) \doteq \bigg\{a\in\mathcal A^+ : \sum_{h\in D, s'\in\mathcal S}\sum_{t=0}^{|h|-1} I_D(h,t,s,a,s') > \tau\bigg\}.
\end{equation}
We estimate the transition probability from a state \(s \in \mathcal S\), to another state \(s' \in \mathcal S\), after taking an admissible action \(a \in \mathcal A(s)\) by dividing the number of times this transition took place by the total number of times the action \(a\) was taken while in state \(s\).
Formally, the count of the number of times the transition took place is defined as $C(s,a,s') \doteq \sum_{h\in D}\sum_{t=0}^{|h|-1} I_D(h,t,s,a,s')$ and the total number of times the action was taken is defined as $C(s,a) \doteq \sum_{s'\in\mathcal S} C(s,a,s')$. 
Thus, we can define an intermediate to the transition function \(\zeta: {\mathcal S\times \mathcal A^+\times \mathcal \mathcal S} \to  [0, 1] \) as $\zeta (s,a,s') = C(s,a,s')/C(s,a)$ for any admissible action $a \in \mathcal A(s)$, and $\zeta (s,a,s')=0$ otherwise.

For the sake of completeness, the ICU-Sepsis environment allows every action \(a\in\mathcal A^+\) in every state \(s\in\mathcal S\) by defining the transition probability distribution of any inadmissible action \(a\notin \mathcal A(s)\) to be the average distribution for all the admissible actions in that state. 
The transition function for the MDP is therefore defined as

\vspace{-15pt}

\begin{equation}
    p(s,a,s') = \begin{cases}
        \begin{aligned}
            &\zeta (s,a,s') \qquad & \text{if } a \in \mathcal A(s) \\
            &\frac{1}{|\mathcal A(s)|} \sum_{a'\in\mathcal A(s)} \zeta (s, a', s') \qquad & \text{if } a \notin \mathcal A(s).
        \end{aligned}
    \end{cases}
\end{equation}

This effectively means that the MDP still only allows the admissible actions to be taken, since taking an inadmissible action is equivalent to choosing one of the admissible actions at random and transitioning accordingly.
Therefore, all optimal policies for the restricted-action setting remain optimal, and all policies that take inadmissible actions in some states can be mapped to equivalent policies that only use admissible actions (by spreading the probability of inadmissible actions across the admissible actions).
This design decision enables the use of RL algorithm implementations that are only compatible with MDPs that allow all actions in all states, without giving them access to inadmissible actions.
We discuss this decision of how inadmissible actions are handled in more detail in Appendix \ref{app:inadmissibleActions}.

An episode ends when the agent reaches the state corresponding to survival or death, after which it can be considered to always transition to $s_\infty$ with probability $1$ regardless of action taken. Therefore, the states corresponding to survival and death are called \emph{terminal states}.

The policy used by the clinicians during the treatment of patients can also be estimated as $\pi_\text{expert} (s,a) \doteq C(s,a)/\sum_{a\in\mathcal A^+} C(s,a)$.
%
%
The initial-state distribution \(d_0\) is defined to be $d_0 (s) \doteq \frac{1}{|D|}\sum_{h\in D}\sum_{a\in \mathcal A^+}\sum_{s'\in \mathcal S} I(h,0,s,a,s')$.
The rewards are determined by the state being transitioned into, with a positive reward (\(R=+1\)) for transitioning into the terminal state corresponding to survival and zero reward for every other transition.

\subsection{Computing the final parameters}\label{subsec:final-params}

The process of clustering the continuous state vectors into a finite set of discrete states (as mentioned in Section \ref{subsec:icu-dataset}) introduces a source of stochasticity in the MDP parameter creation process.
We investigated the effect of different seeds on the resulting MDP by creating 30 environments with different seeds (but which are otherwise identical) and analyzing their properties.
We found that the different environments did not have significantly different properties, so we fixed the seed and defined the resulting MDP to be the ICU-Sepsis MDP.

The result is the transition function \(T\) represented as a tensor of shape \(|\mathcal S| \times |\mathcal A^+| \times |\mathcal S|\), and the reward and initial-state distribution functions vectors \(R\) and \(d_0\), respectively, both represented as vectors of length \(|\mathcal S|\). 
While we have largely followed the work of \citet{komorowski2018the} in the formulation of the MDP, we have made two important changes.
First, the discount factor \(\gamma\) has been set to $1$ instead of $0.99$ to prioritize patient survival over treatment speed.
Secondly, the transition threshold $\tau$ has been increased from $5$ to $20$ to enable more accurate estimation of transition probabilities.
The effects of these changes are examined in Appendix \ref{adx:tau-experiments}.
The values for all the parameters are shown in Table \ref{tab:mdp-params}.

\begin{table}[h]
    \centering
    \begin{tabular}{|ccccc|}
        \hline
        $|\mathcal S|$ & $d_A$ & $n_A$ & $|\mathcal A^+|$ & $\tau$ \\
        \hline
        \nStates & 2 & 5 & 25 & 20 \\
        \hline
    \end{tabular}
    \caption{Parameters for creating the ICU-Sepsis MDP. The values are chosen based on work by \citet{komorowski2018the}, except for $\tau$, where the value has been increased from 5 to 20, to remove actions that are taken very rarely.}
    \label{tab:mdp-params}
\end{table}

\subsection{Additional environment details}\label{subsec:privacy}

The development and release of this environment has prioritized the preservation of patient privacy.
The MDP parameters offer only overarching statistical summaries of patient data, which was previously de-identified during the creation of the MIMIC-III dataset. Consequently, the Institutional Review Board (IRB) review at our institution determined that the MDP and this project are exempt from IRB approval, as the research qualifies as no risk or minimal risk to subjects.
Additionally, the creators of the MIMIC-III dataset affirmed the precedent of model publication derived from the dataset, provided that no straightforward method exists for reconstituting individual patient data.
Therefore, the ICU-Sepsis MDP can be responsibly released, modified, and redistributed for the purposes of RL research without any substantial risk of patient harm.

\begin{table}[h]
    \centering
    \begin{tabular}{| c | ccc | c |}
        \hline
         & Random & Expert & Optimal & Dataset\\
        \hline
        Average return & 0.78 & 0.78 & 0.88 & 0.77\\
        Average episode length & 9.45 & 9.22 & 10.99 & 13.27\\
        \hline
    \end{tabular}
    \caption{Average return and episode lengths for three baseline policies in the ICU-Sepsis MDP---a policy that takes actions uniformly randomly over all actions, the estimated expert policy, and an optimal policy computed by value iteration. The average return and episode lengths in the dataset used to create the ICU-Sepsis MDP are also shown.}
    \label{tab:icu-sepsis-baselines}
\end{table}

Table \ref{tab:icu-sepsis-baselines} shows the baseline properties of the environment and how they compare to the MIMIC-III dataset.
Since the data contains actions selected by trained physicians on real ICU patients, there are relatively few instances of poor decisions in the original dataset. 
This, combined with our removal of actions that were not taken at least $\tau$ times in the dataset, means that the MDP is limited to simulating reasonable treatments. 
If the agent selects poor or unknown treatments (actions that are inadmissible), they are mapped to a uniform distribution over the admissible (i.e., frequently selected) treatments. 
Hence, even an agent that selects actions uniformly randomly achieves a performance similar to that of the expert policy. 
However, the optimal policy computed using value iteration \citep{Bellman:1957} indicates that there is still room for improvement over the expert policy, which can be achieved while only taking actions that clinicians have taken in the real world.

The various design choices involved in the construction of the ICU-Sepsis environment were made with the goal of creating an easy-to-use MDP that is familiar to the RL research community. Notably, while several follow-up works have suggested improvements in the MDP creation process, like time discretization and fluid dose thresholds \citep[see, for example, the work of][]{futoma2020identifying, tang2023leveraging}, we have tried to stay generally faithful to the original design decisions made by \citet{komorowski2018the}.

\section{Experiments}\label{sec:experiments}

The evaluation of RL algorithms often focuses on their ability to learn high-performing policies quickly and reliably. 
Hence, a good benchmark environment is one that not only resembles a real problem of interest, but one that is also challenging enough for modern algorithms that some algorithms are more effective (learn faster, converge to better policies, or learn more robustly) than others. 
To test the ICU-Sepsis MDP, we therefore evaluate several commonly used RL algorithms, including both value function and policy gradient methods, and analyze their learning characteristics.

Specifically, we conducted experiments to answer two research questions: \textbf{1)} How close to optimal are the policies learned using common RL algorithms? 
%
\textbf{2)} How many episodes do common RL algorithms require to find policies that perform nearly optimally?

We conduct experiments using five algorithms that represent a diverse range of approaches commonly used in RL research:  Sarsa \citep{rummery:cuedtr94}, Q-Learning \citep{Watkins1992}, Deep Q-Network \citep{mnih2013playing}, Soft Actor-Critic (SAC) \citep{haarnoja2018soft}, and Proximal Policy Optimization (PPO) \citep{schulman2017proximal}. We use tabular representations for the policies and value functions in all of these algorithms.

\subsection{Methodology}\label{subsec:experiments--methodology}

Hyperparameter tuning is performed using a random search, where each algorithm runs for $300,\!000$ episodes, averaged over eight random seeds for each hyperparameter setting, to maximize expected returns for the last 10\% of the episodes.  After approximating the best set of hyperparameters through the random search, each algorithm is run for $500,\!000$ episodes averaged over $1,\!000$ random seeds to ensure robustness in results. More details about the search and the final hyperparameter values are given in Appendix \ref{adx:hyperparameters-experiments}. We say that an algorithm has \emph{converged} if the average return over the last $1,\!000$ episodes are within $0.1\%$ of the average return over the last $10,\!000$ episodes. 
Since the goal is to find policies with a high expected return in the environment, the returns are not evaluated on a separate MDP built with held-out data, as ICU-Sepsis acts as the ground truth in this case, and generalization of policies to other environments or the real world is not being tested.

\subsection{Results and analysis}

\begin{figure}[h]
     \centering
     \includegraphics[width = \textwidth]{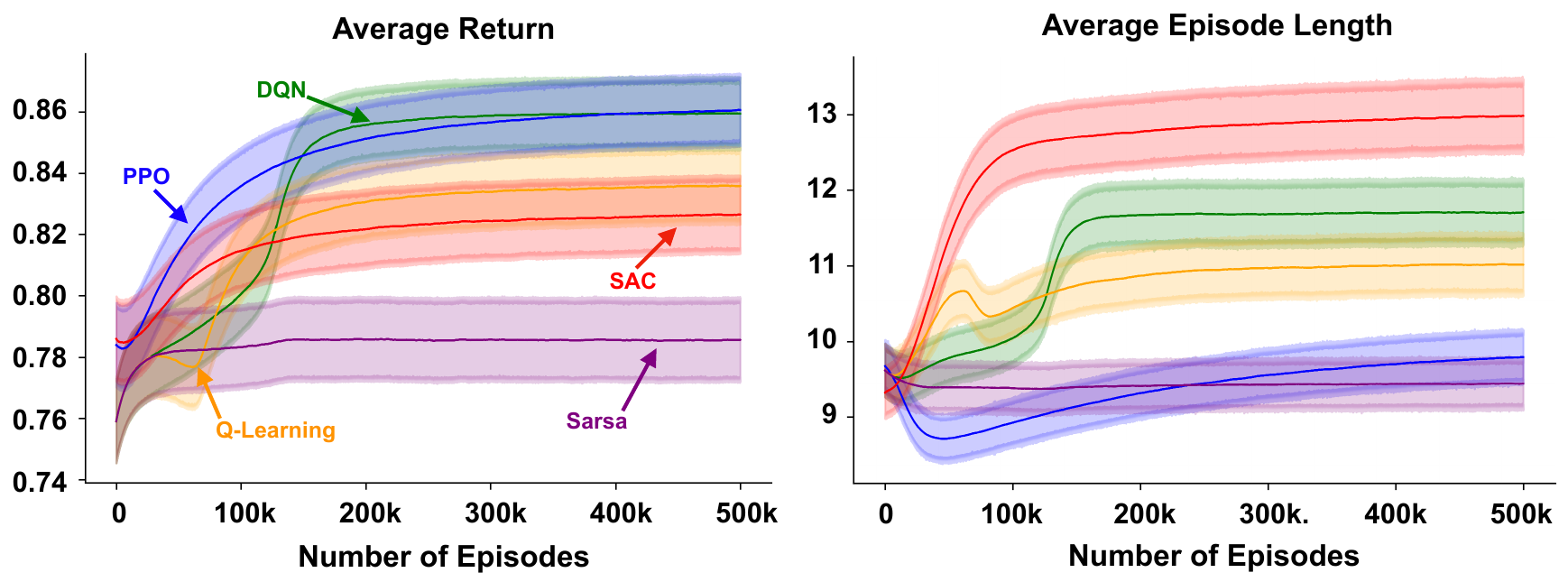}
        \caption{ (\textbf{Left}) The learning curves for five algorithms on the ICU-Sepsis MDP. (\textbf{Right}) Average episode lengths during the learning process. Each curve is averaged over $1,\!000$ random seeds, where the error bars represent one unit of standard error.}
        \label{fig:icu-sepsis-lc}
\end{figure}

Figure \ref{fig:icu-sepsis-lc} shows the learning curves with the average returns and average episode lengths for all five algorithms in the ICU-Sepsis environment. Table \ref{tab:convergence} shows the average number of episodes and time steps needed for each algorithm to converge.


\begin{table}[h]
    \centering
    \begin{tabular}{ r | c c c } \hline 
        \textbf{Algorithm} & \textbf{Episodes (K)} & \textbf{Steps (M)} & \textbf{Average Return} \\ \hline 
        Sarsa & \textbf{105.3}& \textbf{0.99}& 0.79\\
        Q-Learning & 285.8& 3.04& 0.84\\
        Deep Q-Network & 241.5& 2.60& \textbf{0.86}\\
        SAC & 324.0& 4.01& 0.83\\
        PPO & 386.9 & 3.59 & \textbf{0.86} \\ \hline
    \end{tabular}
    \caption{The number of episodes and time steps for each algorithm to converge, as well as the average return over the last $1,\!000$ time steps. It can be observed that the algorithms require a large number of episodes to converge, and not every algorithm is able to achieve near-optimal performance.}
    \label{tab:convergence}
\end{table}

With respect to the first research question, we observe that while some algorithms are able to achieve near-optimal performance, not all algorithms show significant improvement in performance for the learned policy, and notably, the performance of Sarsa is only marginally better than a random agent.
Concerning the second research question, we observe that
%
%
even after extensive parameter tuning, all of these algorithms take hundreds of thousands of episodes (i.e., millions of steps) to converge. 
The average episode lengths are shown in Figure \ref{fig:icu-sepsis-lc} (Right), which are roughly in line with the episode lengths seen in the MIMIC-III dataset, where the episodes had 13.27 steps on average.

\section{Limitations}\label{sec:limitations}

We would like to reiterate that the ICU-Sepsis MDP is designed to model a real-world problem, presenting a level of difficulty for policy search that makes it an excellent environment to evaluate RL algorithms.
However, it is not intended to be a comprehensive medical simulation of sepsis and should not be used for drawing conclusions about treatments for actual patients. 

Sepsis treatment requires careful consideration of numerous factors that are beyond the scope of this MDP.
For example, the vasopressor dosage should change gradually, as abrupt changes can lead to hypertension or cardiac arrhythmia \citep{Fadale2014, Allen2014}, but basing the optimal action solely on the current state may result in policies with numerous sudden changes in vasopressor dosages, deviating from clinically accepted strategies \citep{Jia2020}.
Moreover, the generalizability of the learned policies across different scenarios has not been tested, and these policies might perform suboptimally if treatment standards change over time \citep{Gottesman2019}.

\section{Future Work}

While ICU-Sepsis is designed to be a standardized MDP with broad compatibility with many RL algorithms, it can also serve as the base for another, more medically accurate version of the MDP that incorporates, among others things, the considerations mentioned in Section \ref{sec:limitations}, making it more useful for applied RL research in the healthcare domain.

The choice of creating ICU-Sepsis as a tabular MDP is motivated by the goal of creating an MDP with broad compatibility that also reflects how RL is used in many real-world applications. As mentioned in Section \ref{sec:software}, the normalized values of the state centroids are provided with the MDP, even though the transitions are still modeled in a tabular fashion. However, an additional continuous-state version of the MDP would further broaden the spectrum of RL algorithms that would be suitable to be evaluated on the ICU-Sepsis environment.

\section{Conclusion}

This work introduces the ICU-Sepsis MDP and demonstrates its potential to serve as an environment within benchmarks for RL algorithms.
It is lightweight and easy to set up and use, yet the inherent complexity of the sepsis management task proves to be a significant challenge to modern RL algorithms.
These qualities position the ICU-Sepsis MDP as a strong candidate for inclusion in RL benchmark suites, offering researchers an indicator of the performance of RL algorithms on an important real-world problem.




\bibliography{main}
\bibliographystyle{rlc}

\clearpage

\appendix

\section{Reproducing the ICU-Sepsis Parameters}\label{adx:reproducing}

\begin{figure}[h]
\centering
    \includegraphics[width=0.75\textwidth]{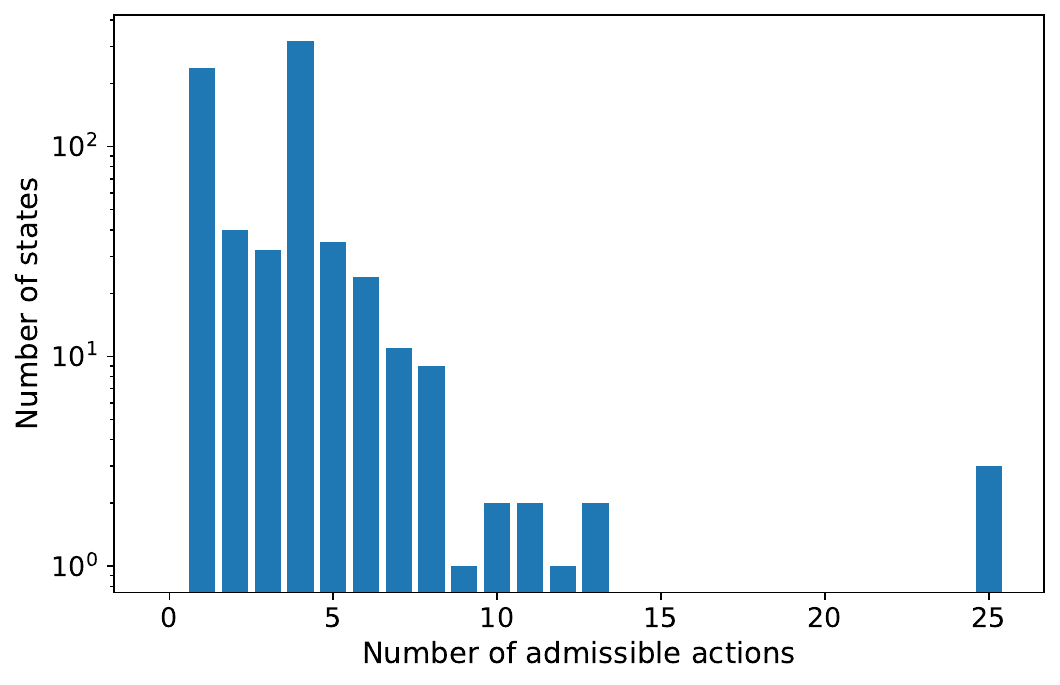}
\caption{Distribution of the number of admissible actions for different states in the ICU-Sepsis environment.}
\label{fig:aic-states}
\end{figure}

The Python code for reproducing the ICU-Sepsis parameters is available in GitHub repository\footnote{\url{https://github.com/icu-sepsis/icu-sepsis}} released with this paper.
Reproducing these parameters would require the researchers to download the MIMIC-III dataset from their website.\footnote{\url{https://physionet.org/content/mimiciii/1.4/}}
The initial steps for identifying patients with sepsis and extracting their features from the MIMIC-III dataset can be performed using the MATLAB code provided in the GitHub repository\footnote{\url{https://github.com/matthieukomorowski/AI\_Clinician}} by \citet{komorowski2018the}.
These steps have also been translated by \citet{Subramanian2020Sepsis} into Python scripts that produce equivalent results with minor differences.
After creating the patient features, estimating the MDP parameters and creating the list of admissible actions can be done using the scripts provided in our GitHub repository. Figure \ref{fig:aic-states} shows the distribution of the number of admissible actions in the states set for ICU-Sepsis.

\section{Handling Inadmissible Actions}
\label{app:inadmissibleActions}

Recall from Section \ref{subsec:const-mdp} that state-action pairs that occur $\tau$ or fewer times in the dataset (where $\tau$ is a hyperparameter) are inadmissible (that is, they cannot be taken) since the subsequent transition distribution is unknown. 
This presents a problem: many implementations of RL algorithms do not allow for different action sets in each state. 
This may present a challenge for researchers hoping to compare to baselines that lack this functionality. 
We therefore opted to design ICU-Sepsis to be compatible with two different perspectives.

In the first perspective, inadmissible actions cannot be taken in the states where they are inadmissible. 
A list of admissible actions for each state is provided in the \texttt{extras/admissibleActions.txt} file provided with the CSV files containing the dynamics, as well as under the \texttt{admissible\_actions} key in the \texttt{info} dictionary provided by the Gym/Gymnasium API. 
Furthermore, the entries in the transition probability table and reward function that correspond to inadmissible state-action pairs can be ignored. 
This perspective is ideal, simulating a setting where inadmissible actions do not exist as options for the agent to consider.

In the second perspective, we ensure that ICU-Sepsis is compatible with software that requires all actions to be admissible in all states. 
A key goal under this perspective is to avoid the fabrication of artificial environment behavior if inadmissible actions are chosen by the agent (e.g., defining inadmissible actions to cause a transition to a state representing death to discourage the selection of inadmissible actions). 
Such artificial transitions are undesirable because they can alter various performance metrics (e.g., performance improvement and learning curve plots could be dominated by the speed with which agents learn not to take inadmissible actions, which is not the important part of the ICU-Sepsis simulation). 
Instead we view inadmissible actions as being truly inadmissible (they cannot be taken by the agent, and hypothetical transitions that result from these actions should not be considered). 
To achieve this, we consider how ICU-Sepsis could be designed so that when RL software selects inadmissible actions, these inadmissible actions are automatically modified to correspond to admissible actions, thereby ensuring that inadmissible actions are never chosen by the agent.

The key insight to enable this is the creation of a mapping from any policy that allows all actions to a corresponding policy that only selects admissible actions. 
Although the agent can learn and reason using a policy that can select all actions, the interactions with the environment (including evaluations of expected return) are equivalent to a corresponding policy that only selects admissible actions.

The most straightforward way to achieve these desired properties would be to define inadmissible actions to instead represent any one of the admissible actions. 
If there is only one inadmissible action, this essentially gives the agent two different ways to select one of the admissible actions. 
Critically, this does not mean that the inadmissible action is actually chosen and the simulated result is the outcome of the inadmissible action. 
Instead, this means that inadmissible actions can never be chosen and instead a redundant policy representation is used (a policy representation that allows for multiple ways of selecting one or more of the admissible actions).

However, this straightforward approach introduces a different issue: in standard RL implementations that require all actions to be allowed in all states, there may not be a mechanism to tell the agent that in some states two different actions actually correspond to a single action. 
When the agent selects one of two equivalent actions, it may not recognize that the outcome of the action provides information about both of the actions. 
That is, the agent will not necessarily generalize properly. 
This raises questions regarding the significance of the choice of \emph{which} action inadmissible actions map to. 
To avoid these complexities, we opt to map inadmissible actions to a distribution over the admissible actions.

Specifically, we define inadmissible actions in a given state to be equivalent to a uniform random selection of the admissible actions in that state. 
This means that if an agent that requires all actions to be allowed in all states selects a inadmissible action, its policy is implicitly modified to uniformly randomly select an action from the admissible set of actions. 
This achieves the desired goals: it is realistic in that it completely disallows actions that it would be irresponsible to allow an RL agent to take (it does not provide hypothetical simulations of the outcomes of these uncertain and risky actions) and it avoids skewing performance metrics because the agent cannot achieve a significant initial increase in expected discounted return by simply learning to avoid inadmissible actions (a uniform random policy over all actions is now equivalent to a uniform random policy over the admissible actions). 
However, it is worth noting the limitation that agents selecting inadmissible actions may still fail to properly generalize, possibly resulting in slower learning than agents that properly handle admissible action sets.

\clearpage

\section{Examining the Effect of the Transition Threshold}\label{adx:tau-experiments}

As explained in Section \ref{subsec:final-params}, the transition threshold has been increased from 5 (as set by \citealt{komorowski2018the}) to 20 to ensure that each admissible action is seen enough times in the dataset to provide a reasonable estimate of the transition probabilities.
To examine the effects of this change on the resulting environment, we create a \emph{Variant} environment with \(\tau=5\) that is otherwise identical to the ICU-Sepsis environment in its creation process, and ask the following research questions about the policies in this new environment: \textbf{1)} What is the highest survival rate possible in the Variant MDP? \textbf{2)} How close to the optimal performance are the policies learned by common RL algorithms? \textbf{3)} How do the average episode lengths change during the learning process for common RL algorithms?

\subsection{Baseline results}

Table \ref{tab:mdp-baseline} shows the baseline results for the Variant MDP and how they compare to ICU-Sepsis. We observe that an optimal policy in the Variant MDP has an expected return of $0.96$, which means that $96\%$ of sepsis patients will survive when treated using this policy, compared to the $77\%$ survival rate seen in the MIMIC-III dataset. Thus, with respect to the first research question, the highest possible survival rate in the Variant MDP appears to be unreasonably high compared to the real data. Table \ref{tab:mdp-baseline--ep-len} also shows that an episode running under this optimal policy will have an expected $24.8$ steps in an episode, much higher that the $13.27$ steps seen in the dataset.

\begin{table}[h]
    \begin{subtable}[h]{0.45\textwidth}
        \centering
        \begin{tabular}{ c c c } \hline 
            \textbf{Agent} & \textbf{ICU-Sepsis} & \textbf{Variant}\\ \hline 
            Random & 0.78 & 0.74 \\
            Expert & 0.78 & 0.77 \\
            Optimal & 0.88 & 0.96 \\ \hline
        \end{tabular}
        \caption{Average return}
        \label{tab:mdp-baseline--perf}
        
    \end{subtable}
    \hfill
    \begin{subtable}[h]{0.45\textwidth}
        \centering
        \begin{tabular}{ c c c } \hline 
            \textbf{Agent} & \textbf{ICU-Sepsis} & \textbf{Variant}\\ \hline 
            Random & 9.45 & 12.6 \\
            Expert & 9.22 & 9.8 \\
            Optimal & 10.99 & 24.8 \\ \hline
        \end{tabular}
        \caption{Average episode lengths}
        \label{tab:mdp-baseline--ep-len}
     \end{subtable}
     \caption{(\textbf{a}) Average return and (\textbf{b}) average episode lengths for ICU-Sepsis and the Variant MDP for three baseline policies: A random policy taking each action uniformly randomly in each state, the expert policy estimated from the dataset, and an optimal policy computed using value iteration. The average return and episode lengths seen in the MIMIC-III dataset were $0.77$ and $13.27$ respectively.}
     \label{tab:mdp-baseline}
\end{table}

\subsection{Performance of various algorithms}

\begin{figure}[h]
     \centering
     \includegraphics[width = \textwidth]{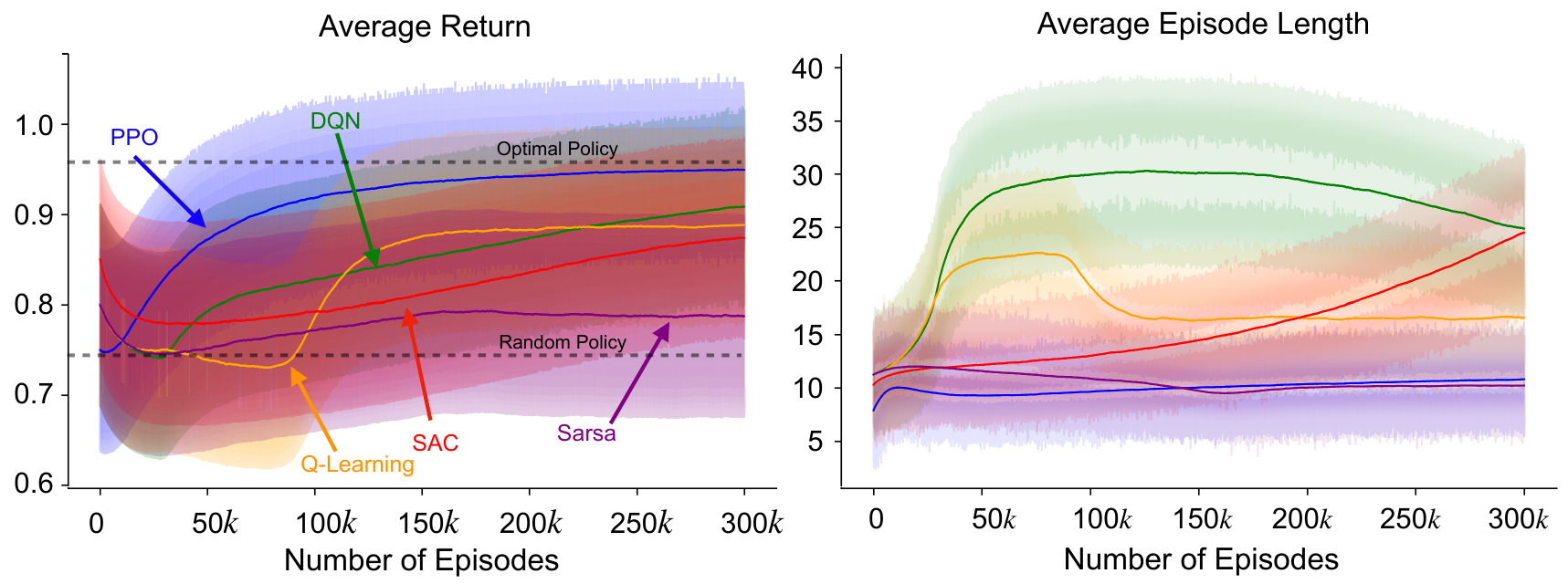}
        \caption{
         (\textbf{Left}) The learning curves for five algorithms on the Variant MDP. (\textbf{Right}) A plot depicting the average episode lengths during the learning process. Each curve is averaged over $20$ random seeds, where the error bars represent one unit of standard error.
        }
        \label{fig:alt-mdp-lc}
\end{figure}

The number of episodes and steps required for convergence and expected returns after convergence are shown in Table \ref{tab:convergence-variant}. Figure \ref{fig:alt-mdp-lc} shows the learning curves and average episode lengths for the five algorithms described in Section \ref{sec:experiments} when run on the Variant MDP, using the same methodology as explained in Section \ref{subsec:experiments--methodology} for the experiments with ICU-Sepsis.

\begin{table}[h]
    \centering
    \begin{tabular}{ r | c c c } \hline 
        \textbf{Algorithm} & \textbf{Episodes (K)} & \textbf{Steps (M)} & \textbf{Average Return} \\ \hline 
        Sarsa & \textbf{125.5} & \textbf{1.42} & 0.79\\
        Q-Learning & 188.3 & 3.48 & 0.89\\
        Deep Q-Network & 283.3 & 7.82 & 0.91\\
        SAC & 273.3 & 4.20 & 0.87\\
        PPO & 235.5 & 2.35 & \textbf{0.95}\\ \hline
    \end{tabular}
    \caption{This table shows the number of episodes and time steps for each algorithm to converge, along with the average return over the last $1,\!000$ time steps.}
    \label{tab:convergence-variant}
\end{table}

Therefore, with respect to the second and third research questions, we see that the expected returns and average episode lengths in the learned policies are unusually high, which do not reflect the numbers seen in the dataset.
We posit that this might be happening because the agent has learned to exploit some rare actions in certain states which happened to result in good outcomes by chance.
Since increasing the transition threshold removes such actions from the set of admissible actions, this behavior is not observed in the ICU-Sepsis MDP which has a higher transition threshold but is otherwise identical in the creation process to the Variant MDP.
To further validate this theory, in Appendix \ref{subsec:perturb} we test the robustness of the three baseline policies: a random policy, the expert policy, and the optimal policy learned using value iteration for both ICU-Sepsis and the Variant.

\subsection{Effect of perturbations on the environments}\label{subsec:perturb}

To illustrate the robustness of different policies in the ICU-Sepsis and the Variant MDP, we evaluate the performance of the baseline policies after making some perturbations in the environment dynamics.
Each environment (ICU-Sepsis and the Variant) is first perturbed in the following way:

\begin{enumerate}[itemsep=0.5em, parsep=0em, topsep=0em]
    \item Among all the admissible actions, each of them is made inadmissible with some probability \(\sigma \in [0,1]\) independently of each other.
    \item If all of the actions for some state are made inadmissible, one of the previously admissible actions for that state is randomly chosen and made admissible again. Thus, every state will always have at least one admissible action.
    \item As explained in Section \ref{subsec:const-mdp}, any inadmissible action taken by the agent is equivalent to randomly choosing one of the admissible actions (according to the new list of admissible actions after the perturbation process) and taking that action.
\end{enumerate}

\begin{figure}[t]
     \centering
     \includegraphics[width=\textwidth]{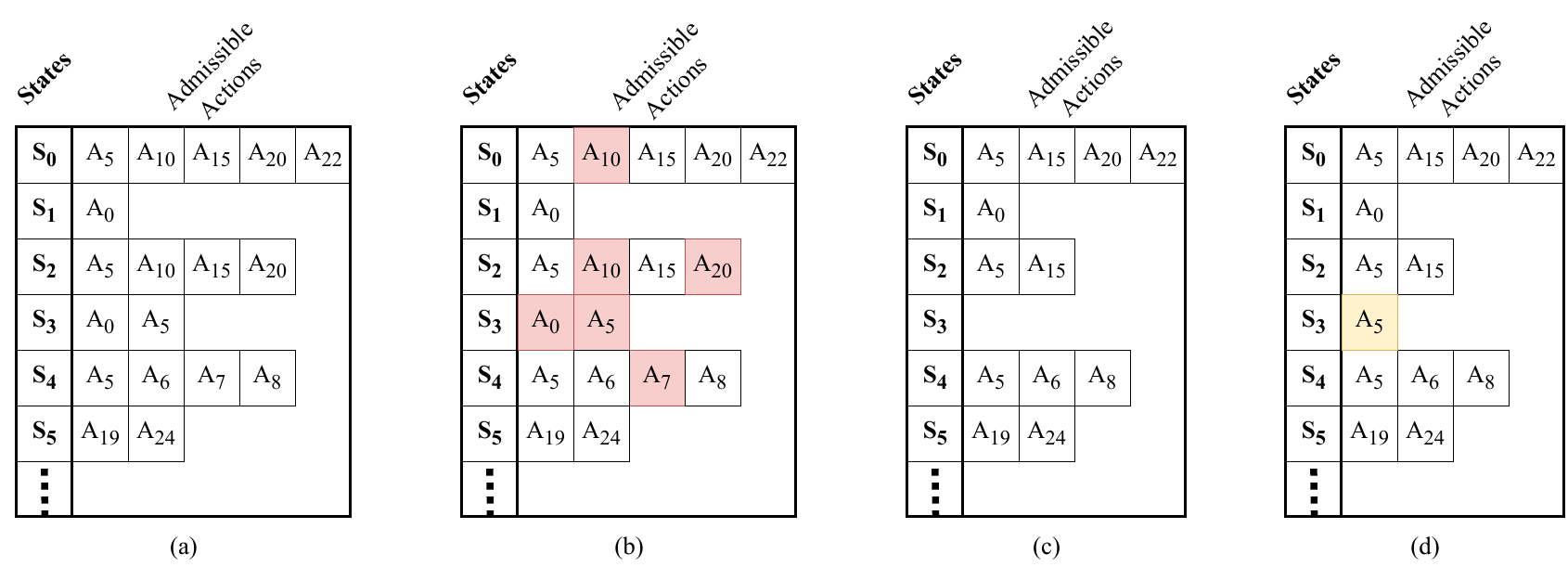}
     \caption{Illustration of the perturbation process. \textbf{(a)} Admissible actions for different states. Each row has a state (in bold) followed by the list of admissible actions in that state. \textbf{(b)} Some admissible actions are randomly chosen and made inadmissible. \textbf{(c)} Remaining admissible actions. This can cause some states (in this case $\text{S}_3$) to have no admissible actions left. \textbf{(d)} For states where there are no admissible actions left, a previously admissible action is chosen and reintroduced as an admissible action. Thus, every state still has at least one admissible action after the perturbation process.}
     \label{fig:perturb-illustration}
\end{figure}

\begin{figure}[t!]
     \centering
     \begin{subfigure}[b]{\textwidth}
         \centering
         \includegraphics[width=\textwidth]{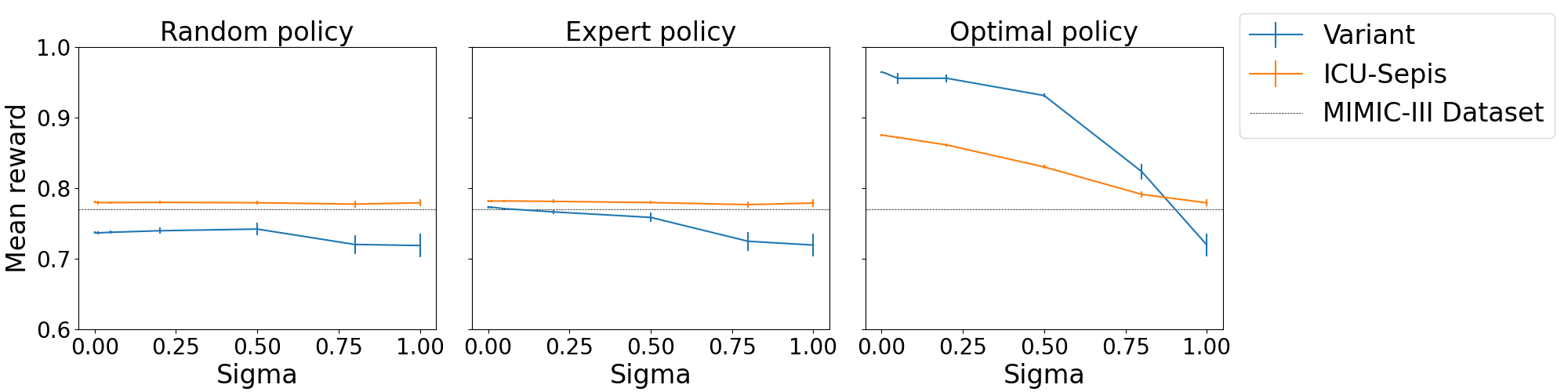}
         \caption{Return}
         \label{fig:perturb-rew-1}
     \end{subfigure}

     \begin{subfigure}[b]{\textwidth}
         \centering
         \includegraphics[width=\textwidth]{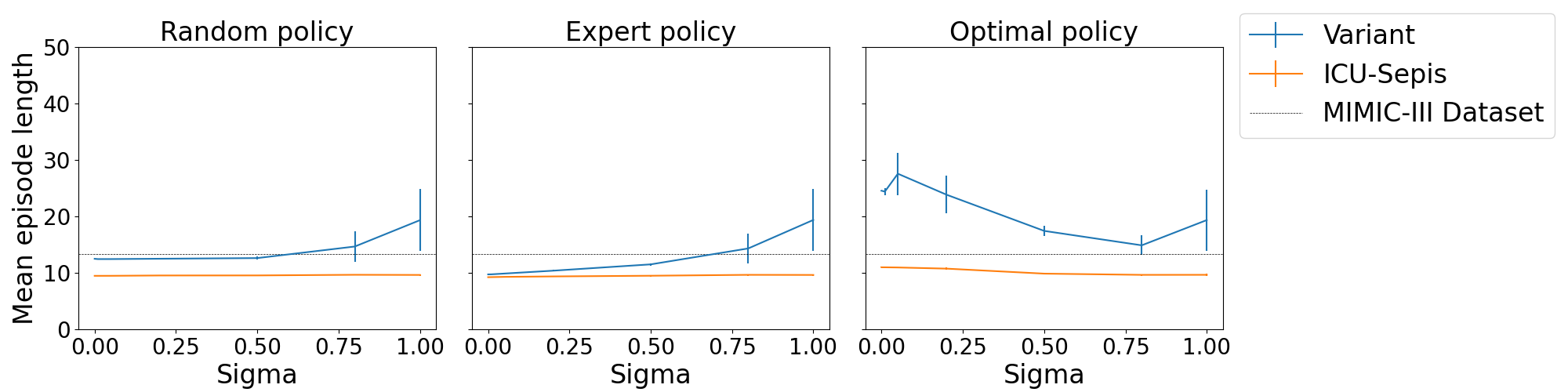}
         \caption{Number of steps per episode}
         \label{fig:perturb-ep-len-1}
     \end{subfigure}
        \caption{Effects of removing some actions from the set of admissible actions on the learned policies as the probability of removing actions (\(\sigma\)) increases from $0$ to $1$. Each perturbation was done 32 times for each environment and the average and standard error of the results are shown. (\textbf{a}) The average return for different policies. (\textbf{b}) The average lengths of episodes for different policies.}
        \label{fig:perturb-results-1}
\end{figure}

Figure \ref{fig:perturb-illustration} shows an illustration of this process, which is repeated 32 times for each policy in each environment. 
%
%
If a policy is over-reliant on a few transitions, then their removal should result in a large performance drop. Therefore, such policies should have higher variance across runs, where some runs would not allow the actions that are being exploited to obtain unrealistically high returns.
Figure \ref{fig:perturb-results-1} shows that the variance is indeed higher for the Variant compared to ICU-Sepsis, where the average return and episode lengths stay more stable as actions are progressively made inadmissible.

\clearpage

\section{Hyperparameters Search}\label{adx:hyperparameters-experiments}

The algorithms have been implemented as modifications on top of the CleanRL\footnote{\url{https://github.com/vwxyzjn/cleanrl}} library \citep{huang2022cleanrl}. 

\subsection{Random search setting}

\begin{table}[h]
    \centering
    \begin{tabular}{ l | l | l }
        \hline
        \textbf{Hyperparameter} & \textbf{Value(s) / Range} & \textbf{Distribution }\\
        \hline
        Number of Seeds & 8 & - \\
        Learning Rate & $[10^{-5}, 0.01]$ & Log Uniform \\
        Number of Environments & 1 & - \\
        Buffer Size & $[1^3, 10^6]$ & Integer Log Uniform \\
        Discount Factor ($\gamma$) & 1.0 & - \\
        Polyak Averaging Coefficient ($\tau$) & $[0.001, 1.0]$ & Log Uniform \\
        Target Network Update Frequency & $[1, 1000]$ & Integer Uniform \\
        Batch Size & $[1, 256]$ & Integer Uniform \\
        Start Exploration Rate ($\epsilon_{start}$) & $[0.01, 1.0]$ & Uniform \\
        End Exploration Rate ($\epsilon_{end}$) & $[0.01, 0.1]$ & Log Uniform \\
        Exploration Fraction & $[0.0, 1.0]$ & Uniform \\
        Learning Starts & 10,000 & - \\
        Training Frequency & 10 & - \\
        \hline
    \end{tabular}
    \caption{Hyperparameter settings and distribution types for the DQN hyperparameter search.}
    \label{tab:dqn-hyperparameters}
\end{table}

\begin{table}[h]
    \centering
    \begin{tabular}{ l | l | l }
        \hline
        \textbf{Hyperparameter} & \textbf{Value(s) / Range} & \textbf{Distribution }\\
        \hline
        Number of Seeds & 8 & - \\
        Learning Rate & $[10^{-5}, 0.01]$ & Log Uniform \\
        Number of Environments & 1 & - \\
        Number of Steps & $[100, 500]$ & Integer Uniform \\
        Number of Mini-batches & $[1, 6]$ & Integer Uniform \\
        Discount Factor ($\gamma$) & 1.0 & - \\
        GAE Lambda & $[0.0, 1.0]$ & Uniform \\
        Update Epochs & $[1, 8]$ & Integer Uniform \\
        Normalize Advantage & True & - \\
        Clipping Coefficient & $[0.1, 0.5]$ & Uniform \\
        Clip Value Loss & True/False & - \\
        Entropy Coefficient & $[10^{-2}, 1.0]$ & Log Uniform \\
        Value Function Coefficient & $[0.2, 1.0]$ & Uniform \\
        Maximum Gradient Norm & $[0.1, 1.0]$ & Uniform \\
        Target KL &  $[ \text{Null }, 0.01, 0.05, 0.1]$ & Uniform \\
        \hline
    \end{tabular}
    \caption{Hyperparameter settings and distribution types for the PPO hyperparameter search.}
    \label{tab:ppo-hyperparameters}
\end{table}

\begin{table}[ht]
    \centering
    \begin{tabular}{ l | l | l }
        \hline
        \textbf{Hyperparameter} & \textbf{Value(s) / Range} & \textbf{Distribution }\\
        \hline
        Number of Seeds & 8 & - \\
        Learning Rate & $[10^{-5}, 0.01]$ & Log Uniform \\
        Number of Environments & 1 & - \\
        Buffer Size & 1 & - \\
        Discount Factor ($\gamma$) & 1.0 & - \\
        Batch Size & 1 & - \\
        Start Exploration Rate ($\epsilon_{start}$) & $[0.01, 1.0]$ & Uniform \\
        End Exploration Rate ($\epsilon_{end}$) & $[0.01, 0.1]$ & Log Uniform \\
        Exploration Fraction & $[0.0, 1.0]$ & Uniform \\
        \hline
    \end{tabular}
    \caption{Hyperparameter settings and distribution types for the Q-learning and Sarsa hyperparameter search.}
    \label{tab:qlearning-hyperparameters}
\end{table}

\begin{table}[ht]
    \centering
    \begin{tabular}{ l | l | l }
        \hline
        \textbf{Hyperparameter} & \textbf{Value(s) / Range} & \textbf{Distribution }\\
        \hline
        Number of Seeds & 8 & - \\
        Buffer Size & $[10^3, 10^6]$ & Integer Log Uniform \\
        Polyak Averaging Coefficient ($\tau$) & $[10^{-3}, 1.0]$ & Log Uniform \\
        Batch Size & $[1, 256]$ & Integer Uniform \\
        Learning Starts & $[10^4, 2 \times 10^4]$ & - \\
        Policy Learning Rate & $[10^{-5}, 0.01]$ & Log Uniform \\
        Q-function Learning Rate & $[10^{-5}, 0.01]$ & Log Uniform \\
        Update Frequency & $[1, 6]$ & Integer Uniform \\
        Target Network Update Frequency & $[100, 10^4]$ & Integer Uniform \\
        Temperature Coefficient ($\alpha$) & $[0.01, 1.0]$ & Uniform \\
        Automatic Entropy Tuning & False/True & - \\
        Target Entropy Scale & $[0.01, 1.0]$ & Uniform \\
        Number of Environments & 1 & - \\
        Discount Factor ($\gamma$) & 1.0 & - \\
        \hline
    \end{tabular}
    \caption{Hyperparameter settings and distribution types for the SAC hyperparameter search.}
    \label{tab:sac-hyperparameters}
\end{table}

Weights \& Biases (Wandb)\footnote{\url{https://wandb.ai/}} \citep{wandb} was utilized for performing the random search over hyperparameters. The ranges and distributions used for the searches across different algorithms are detailed in Tables \ref{tab:dqn-hyperparameters}, \ref{tab:ppo-hyperparameters}, \ref{tab:qlearning-hyperparameters}, and \ref{tab:sac-hyperparameters}. To ensure equitable compute resources across different methods, each was allocated 72 CPUs and a maximum duration of 4 days for the search, with the process concluding at that time. The number of hyperparameters explored for each method is listed in Table \ref{tab:method-runs}, highlighting that slower methods were limited to fewer parameter searches. Altogether, $\geq 11,\!000$ parameters were searched across all methods.

\begin{table}[ht]
    \centering
    \begin{tabular}{ c | c }
        \hline
        \textbf{Method Name} & \textbf{Number of Hyperparameters} \\
        \hline
        Q Learning & 2263 \\
        Sarsa & 2501 \\
        SAC & 1162 \\
        PPO & 3224 \\
        DQN & 2632 \\
        \hline
    \end{tabular}
    \caption{Number of runs for different methods.}
    \label{tab:method-runs}
\end{table}

\clearpage

\subsection{Best set of approximated hyperparameters}
Table \ref{tab:hyperparameters_comparison} lists the best hyperparameters for each method found during the random search. These hyperparameters were used in the experiments, with results shown in Figure \ref{fig:icu-sepsis-lc}.

\begin{table}[h]
    \centering
    \begin{tabular}{ l | c | c | c | c | c } \hline
        \textbf{Hyper-parameter} & \textbf{DQN} & \textbf{PPO} & \textbf{Q-Learning} & \textbf{SAC} & \textbf{Sarsa} \\ \hline
        Learning Rate & 0.001 & 0.005 & 0.0025 & $\pi$: 0.025, Q: 0.025 & 0.0025 \\
        Optimizer & Adam & Adam & Adam & Adam & Adam \\
        Buffer Size & 10,000 & & 1 & 10,000 & 1 \\
        Batch Size & 64 & & 1 & 64 & 1 \\
        Start Exploration Rate ($\epsilon$ start) & 1.0 & & 1.0 & & 1.0 \\
        End Exploration Rate ($\epsilon$ end) & 0.001 & & 0.001 & & 0.001 \\
        Exploration Fraction & 0.25 & & 0.1 & & 0.25 \\
        Learning Starts & 10,000 & & & 10,000 & \\
        Training Frequency & 10 & & & & \\
        Number of Steps for Rollout & & 500 & & & \\
        Number of Minibatches & & 1 & & & \\
        GAE Lambda & & 0.4 & & & \\
        Update Epochs & & 6 & & & \\
        Normalize Advantage & & Yes & & & \\
        Clipping Coefficient & & 0.5 & & & \\
        Clip Value Loss & & No & & & \\
        Entropy Coefficient & & 0.005 & & & \\
        Value Function Coefficient & & 0.3 & & & \\
        Maximum Gradient Norm & & 0.4 & & & \\
        Target KL Divergence & & 0.001 & & & \\
        Polyak Average ($\tau$) & 0.01 & & & 0.01 & \\
        Target Network Update Frequency & 512 & & & 500 & \\
        Update Frequency & & & & 1 & \\
        Alpha & & & & 0.25 & \\
        Autotune & & & & No & \\
        Target Entropy Scale & & & & 0.2 & \\ \hline
    \end{tabular}
    \caption{ Hyper-parameters used in DQN, PPO, Q-Learning, SAC, and Sarsa  to solve ICU-Sepsis.}
    \label{tab:hyperparameters_comparison}
\end{table}


\end{document}